\documentclass{article}
\usepackage{spconf,amsmath,epsfig,xcolor}
\usepackage{pgfplots}
\pgfplotsset{compat=1.17}

\usepackage{graphicx}
\usepackage{caption,subcaption}
\usepackage{hyperref}

\usepackage{tikz}

\usepackage{enumitem}
\usepackage{times}
\usepackage{epsfig}
\usepackage{graphicx}
\usepackage{amsmath}
\usepackage{amssymb}
\usepackage{booktabs}
\usepackage{float}
\usepackage{xspace}
\usepackage{pifont}
\usepackage{siunitx}
\usepackage{xcolor}
\usepackage[normalem]{ ulem }
\usepackage{soul}
\usepackage{array}
\usepackage{multirow} 
\usepackage[export]{adjustbox}
\usepackage{tabularray}

\title{Leveraging edge detection and neural networks for better UAV
localization}

\name{
	\begin{tabular}{ccc}
		Théo Di Piazza$^1$&
		Enric Meinhardt-Llopis$^2$&
		Gabriele Facciolo$^2$\\
		Bénédicte Bascle$^1$&
		Corentin Abgrall$^1$&
		Jean-Clément Devaux$^1$
	\end{tabular}
}

\address{
	$^1$Thales LAS, France\hspace{0.5cm}%
	$^2$Université Paris-Saclay, CNRS, ENS Paris-Saclay, Centre Borelli, France}

\begin{document}


\maketitle

\begin{abstract}
	We propose a new method for the geolocalization of Unmaned Aerial
	Vehicles (UAV) in environments without Global Navigation Stallite
	Systems (GNSS).
	Current state-of-the-art methods use an offline-trained encoder to
	compute a vector representation (embedding) of the current UAV's view,
	and compare it with the pre-computed embeddings of geo-referenced
	images in order to deduce the UAV's position.
	Here, we show that the performance of these methods can be greatly
	improved by pre-processing the images by extracting their edges, which
	are robust to seasonal and illumination changes. Moreover, we also
	show that using edges  improves the robustness to orientation and
	altitude errors.  Finally,  we present a confidence criterion for
	localization.  Our findings are validated using synthetic experiments. The code is available at \href{https://github.com/theodpzz/uav-localization}{\color{blue}{ https://github.com/theodpzz/uav-localization}}.


\end{abstract}

\section{Introduction}

Unmanned Aerial Vehicles (UAVs) usually accomplish autonomous navigation
through accurate global navigation satellite systems (GNSS). However, when GNSS
systems are not accessible due to loss of signal or jamming for example, the
UAV's camera can be used to help locate it.  Visual Odometry (VO)~\cite{vo1}
estimates the camera displacement between successive images and is often used
for UAV navigation. However, the accuracy of VO relies on the ability to
extract reliable features across frames, which can be difficult in environments
with little texture or repetitive patterns such as  forests or uniform plains.
In addition, VO accumulates drift with time, and this drift needs correction at
regular time intervals. As an alternative to VO, Simultaneous Localization And
Mapping (SLAM)~\cite{slam1} builds a map of the environment that
the camera is moving within. However, both of these methods are usually applied in
indoor environments, with short trajectories or altitudes very close to the
ground \cite{bianchi, slam1, weiss}.  Visual SLAM presents as much drift as VO
if no loop closure or revisit constraints are used.


\begin{figure}[t]
    \centering

\vspace{-0.5em}

        \includegraphics[width=1\columnwidth]{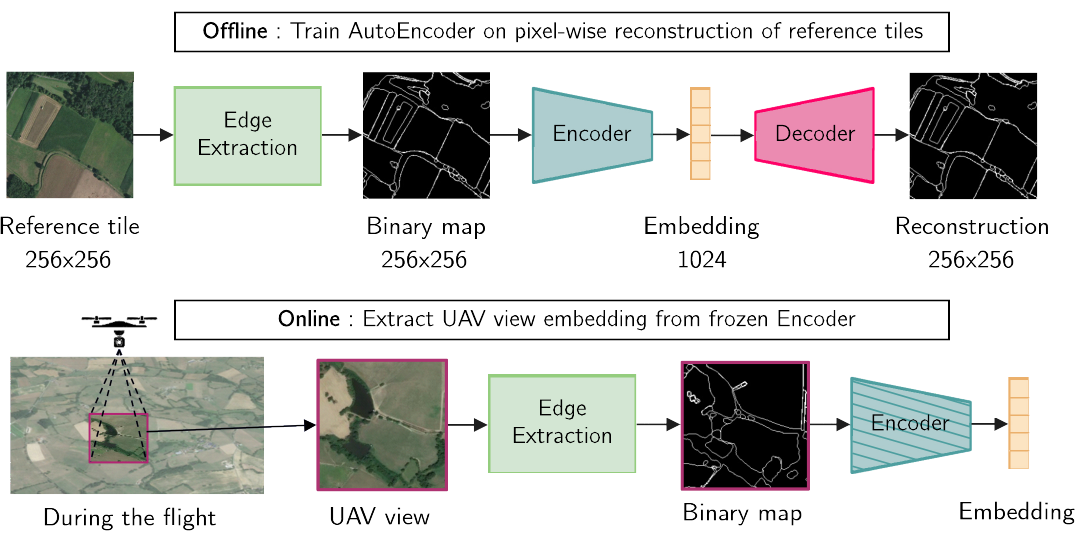}

\vspace{-0.5em}
        
        \vspace{.75em}
        
        \caption{\small Overview of the method for edge extraction and Auto-Encoder. Offline, the Auto-Encoder is trained to reconstruct edges of the tiles from the reference area. Then embeddings of reference tiles and weights of the encoder are loaded on board the UAV. During the UAV flight, edges are extracted from the UAV view before being given to the encoder which produces an embedding. Last, the UAV's position is estimated by comparing the embedding of the UAV's view with the embeddings of the reference tiles.}
        \label{fig:teaser}
\end{figure}

Taking advantage of the easy access to large databases of geo-referenced
satellite images, some recent methods propose to compare the UAV's view with a
set of geo-referenced satellite images. Patel et al.~\cite{patel} propose to
localize the UAV by comparing, via mutual information, the UAV view with
reference images acquired around the flight path.
However, this method requires to store a large quantity of images on board the
UAV.
To circumvent this issue, Bianchi and Barfoot~\cite{bianchi} propose to compare
lightweight vector representations (embeddings) of the UAV's views with
reference images, rather than comparing the whole images. These embeddings are
obtained using an Auto-Encoder trained to reconstruct, pixel by pixel, images
of the area around the trajectory. In a similar vein, Couturier and
Akhloufi~\cite{couturier} propose to use a triplet model to train an encoder so
that nearby areas (that are visually similar) are also close in the embedding
space. The authors claim that this property would allow to interpolate nearby
embeddings in order to refine the position of the UAV. However, the training
loss does not enforce the interpolation property, and as we will see in our
experiment, this strategy does not provide any additional precision.

Note that these methods must define a reference area around the planned
trajectory in which the UAV should remain in the event of signal loss.  In
essence they aim to learn a bijective mapping from a set of images to an
embedding space. But the embedding should be invariant to small appearance
changes and small geometric deformations.

The literature also addresses the issue of matching ambiguity and robustness to
deformation in a limited manner. Bianchi and Barfoot~\cite{bianchi} study the
influence of lighting conditions. They assume that roll, pitch and altitude are
known. Kinnari et al.~\cite{kinnari} focus on season-invariant localization.
They train for given altitude and pitch.

In this work we focus on embedding-based approaches, but instead of working
directly with grayscale images, we propose to use an edge extraction step
before encoding the images (see Figure~\ref{fig:teaser}).  Since image edges
are geometric features, they are very robust to large variations of color and
illumination between satellite and aerial images taken at different dates.
Thus, the method is natively robust to changes that do not even appear in the
training database.
This significantly increases localization performance and improves the
robustness of the model.
Last, a confidence criterion for location prediction is presented aimed at
reducing the number of wrong locations.


\section{Related work}

\smallskip \noindent \textbf{Aerial Visual Localization}  \label{aerial}
Conte and Doherty~\cite{conte} propose an image matching method between the UAV view and the nearest georeferenced image based on Sobel edge extraction, but without any encoders. This method has the advantage of being efficient at high altitudes, as it benefits from the ability to extract structures from edges. 
In~\cite{majdik} the authors manage to locate a quadroctoper in an urban environment by comparing the quadroctoper's view with a set of images available on Google Street View. More recently, Patel et al.~\cite{patel} 
proposed a method for matching the UAV view image to reference images using mutual information, which makes it robust to differences in light conditions. Similarly, the authors of~\cite{gurgu2022vision} use deep features (with Superglue~\cite{sarlin2020superglue}) for performing the matching. However, these methods have to carry all the reference images on board the UAV.
 
\smallskip \noindent \textbf{Bag of Visual Words}
Positioning the UAV by identifying the nearest reference image is akin to an image retrieval task. 
Bag of Visual Words (BoVW)~\cite{bovw} is a method that enables to represent an image as a bag of its visual features that can be extracted by classic SIFT~\cite{sift} features or modern variants as SuperPoints~\cite{superpoint}.
%
BoVW can be used to find the  reference image (and its localization) that best match the representation vector associated to the current UAV's view. 


The BoVW method gathers features from all the images and group them into  $k$ clusters. For a given image, the representation (embedding) is the  a frequency vector that records how many features (from the image) fall into each cluster.
Similarity scores are then computed using a scalar product between embeddings. 
\smallskip \noindent \textbf{Auto-Encoder}
Bianchi and Barfoot~\cite{bianchi} propose to use an Encoder to project the reference and UAV view images into a latent space of smaller dimension. Since the latent embedding is much smaller than the images this   provides  a solution to the onboard storage constraint of the UAV.
Offline, an Auto-Encoder (encoder $E$, and decoder $D$ pair) is trained to reconstruct pixel-by-pixel the reference images $I$. The L2 loss is used for this training
\begin{equation}
L_{AutoEncoder} = \|I - D(E(I))\|_{2}^{2},
\end{equation}
using images of $256 \times 256$ pixels.
Only the embeddings of the reference images $E(I)$ and encoder weights are stored on board the UAV. Online, the encoder $E$ computes the embedding from the UAV's view, and estimates its position by a weighted average of the positions of all the reference images weighted by the similarity to the embedding current UAV view.


\smallskip \noindent \textbf{Triplet Model}
The Triplet Model introduced in~\cite{facenet} presents an alternative way for training an encoder.
Couturier and Akhloufi~\cite{couturier}  apply it to the context of UAV localization. 
The encoder is trained to minimize the distance between embeddings of the same scene and maximizing it for embeddings of different scenes. During training, triplets of images are sampled: an Anchor  
$I_{a}$ and a Positive image captured at the same position  but at different instants; and a Negative image $I_{n}$ captured at the same instant as the Anchor, but at a different position. For each image the encoder produces an embedding $x_{a}$, $x_{p}$ and $x_{n}$. A triplet loss is used to force $x_{a}$ to be closer to $x_{p}$, and $x_{a}$ further away from $x_{n}$
\begin{equation}
    L_{Triplet} = \max(0, \|x_{a} - x_{p}\|_{2}^{2} - \|x_{a} - x_{n}\|_{2}^{2} + \alpha),
\end{equation}
where  $\alpha$  is the margin that keeps negative samples far apart.

%

\begin{figure}[t]
\begin{center}
\includegraphics[width=1.0\linewidth]{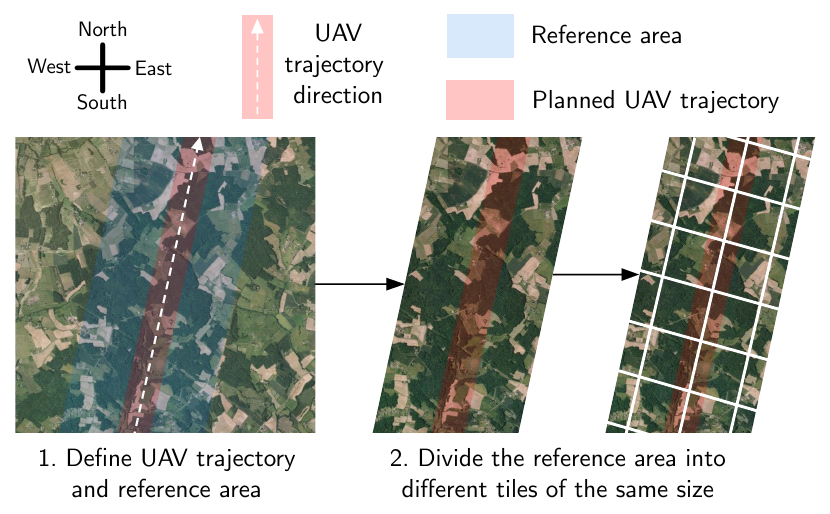}
\end{center}
        \vspace{-2em}

   \caption{\small Illustration of the process for generating the reference tiles from imagery of the reference area.}
\label{fig:refarea}
\end{figure}

\begin{table}
    \centering
    \small
    \begin{tabular}{c|cccc}
         Trajectories & A & B & C & D \\
         \hline
         Altitude (m) & 40 & 2000 & 4000 & 2000 \\
         Distance (m) & 150 & 12000 & 50000 & 2000 \\
         Search band width (m) & 10 & 1000 & 5000 & 2000
    \end{tabular}
    \caption{\small Details about trajectories A, B, C and D. Reference tiles are generated from IGN BD Ortho  for a given year. UAV views are generated from images for a different year. There is 99.5\% overlap between 2 adjacent tiles for trajectory A, and 87.5\% for trajectories B, C and D.}
    \vspace{-1em}
    \label{tab:database}
\end{table}

\section{Dataset}

We performed our experiments on simulated UAV trajectories extrated from real aerial imagery from the  publicly available aerial survey by the French \emph{Institut Géographique National} (IGN) called \emph{DB Ortho}~\cite{IGNdbortho}. This collection of orthophotographs  of the French territory is captured at 20 cm resolution, and updated every 3-4 years. Hence, for each location,  images from  different years are available. 

We defined 4 trajectories A, B, C and D with the characteristics listed in Table~\ref{tab:database}, for each trajectory a reference area in which the UAV should stay is also defined. 
As illustrated in Figure~\ref{fig:refarea}, the reference area is divided into tiles (256x256) representing a part of this reference area. The training set is made of images of the reference area in a given year, while the testing set is made up of images associated with the UAV view along the planned trajectory in a different year.

In our first set of experiments, altitude and UAV orientation are supposed
known (by inertial measurements). In the second part of our experiments, we
investigate the influence of errors in altitude and orientation.




\section{Methodology}


Two images of the same area taken at different dates can be very different due
to seasonal and anthropogenic factors: change in the color of a field, decrease
in greenery, destruction of forest, new buildings etc.  However, over
relatively long periods, much of the geometric visual information remains
unchanged: roads, enclosures, forest delimitation and buildings. This
information is present in the image edges and is independent of pixel
intensities. Thus, applying an edge detector to all the images should improve
the localization robustness.
Here we consider two edge detection methods:
\vspace{-0.5\topsep}
\begin{itemize}[leftmargin=*]
\setlength\itemsep{0em}
    \item 
The classic Canny edge detector~\cite{canny}, which identifies
significant variations in image intensity that correspond to contours.
Given an image it extracts a binary map of these contours.
We investigated different parameter choices and the best results are obtained
with a threshold in the interval [100, 200] and a kernel of size 3.

    \item 
Edges computed using the 
Segment Anything Model~\cite{sam} (SAM), which is a Vision Transformer (ViT) model
that can ‘‘\textit{cut out any object in any image}’’. We apply SAM to each
image to obtain a segmentation masks, which are then transformed into  binary
maps containing all the segmentation boundaries. SAM is available in different
mode sizes: Base, Large and Huge. 
We use SAM Base as it achieves equivalent localization results to the larger
models, for significantly less time inference.

\end{itemize}

 \smallskip

\noindent \textbf{Considered methods} 
To assess the contribution of the edge extraction, we first establish a baseline using Bag of Visual Words computed on grayscale images. Then  Auto-Encoder~\cite{bianchi} and a Triplet Model~\cite{couturier} will be tested on grayscale and on edges extracted using Canny and SAM. 

In the present work, we use the position of the reference embedding leading to the highest scalar product~\cite{bianchi} as we observed that (in our setting) it yields better results than the weighted average. The details of the three methods used are: 
\vspace{-0.5\topsep}
\begin{itemize}[leftmargin=*]
\setlength\itemsep{0em}
    \item 
{Bag of Visual Words}. The number of clusters $k$ tried ranges from $256$ to $10 000$. For visual descriptor extraction, we tested SIFT~\cite{sift} and SuperPoint~\cite{superpoint}. In the results category, the best performances will be displayed.

\item {Auto-Encoder}. The Encoder consists of 4 strided convolutional layers, followed by batch normalization and a LeakyReLU activation function. The latent space dimension is 1024. The Decoder consists of 4 deconvolution layers, followed by batch normalization and a LeakyReLU activation function. If the inputs are binary maps, a sigmoid is added to the last layer of the Decoder. The neural network has 172 506 561 parameters. 
The model was trained for 200 epochs on batches of size 32 using AdamW with an initial learning rate of 0.0003. The \emph{ReduceLROnPlateau} scheduler was used with a factor of $0.5$. 

\item {Triplet Model}. The Encoder architecture corresponds to the ResNet34 \cite{resnet} implementation. The latent space dimension is 1024. The neural network has 21 278 400 parameters. We add a hard negative mining step to our implementation, which significantly increases localization performance.
The model was trained for 200 epochs on batches of 512 using the Adam optimizer with an initial learning rate of 0.001 and  batches of 512. 

\end{itemize}

\begin{table}[t]
\small
\centering
\begin{tabular}{c c | c | c | c | c }
\toprule
\multicolumn{2}{c}{Trajectory ID} & A & B & C &  D\\
\toprule
BoVW & Grayscale  & 50.40 &  23.92 &  2.67  & 1.20 \\
\midrule
Triplet & Grayscale   & 16.02  & 82.16 & 78.60 &  54.21 \\
{}  & Canny   & 43.91 &  67.94 & 85.28& 44.58 \\
{} & SAM   & 12.42  & 30.06  & 78.26 & 67.47 \\
\midrule
Auto-Encoder & Grayscale  & 86.24  & 71.20  & 92.00  & 50.00 \\
{}  & Canny   & 88.16 &  90.20 &  \textbf{99.67} & \textbf{88.10}\\
{}  & SAM   & \textbf{89.76} & \textbf{90.40} & 98.00  & 83.33 \\
\bottomrule
\end{tabular}
\caption{\small Comparison of localization accuracy at 15m for different methods and representations.}
\label{table:results}
\end{table}

\begin{table}[t]
\small
\centering
\centering
\begin{tabular}{c | c | c | c | c}
\toprule

Traj. ID & A &  B &  C &  D\\ 
\toprule

 Graysc.  & 0 / 0 & 89.5 / 72.8 & 99.2 / 91.3 & 82.0 / 46.4\\
 Canny   & \textbf{96.0} / \textbf{12.1} & \textbf{99.7} / \textbf{91.0} & 98.6 / 85.7 & \textbf{100} / \textbf{79.7}\\
  SAM   & 95.0 / 0.2 & 96.4 / 89.8 & \textbf{99.3} / \textbf{98.6} & 100 / 67.8\\
\bottomrule
\end{tabular}

\caption{\small Results \underline{for the Auto-Encoder method} using the confidence criterion.
For each trajectory, value pairs  (\textit{acc.}/\textit{\%}) correspond  to: the percentage of UAV views correctly located at 15m, and to the percentage of UAV views not rejected by the confidence criterion. The  threshold for the confidence criterion was set to $1.13$. Note that the accuracy does not count the predictions rejected by the criterion.}
\label{table:results_lowe}
\end{table}

\smallskip 

\noindent \textbf{Confidence criterion} \label{conf}
The location of the UAV computed by the proposed method corresponds to the reference image with the highest similarity score, but this ignores the similarity scores of the other reference images, which could help defining a confidence criterion. 
%
Thus, we propose to adopt a confidence criterion based on the Lowe's Ratio~\cite{sift,scharstein1998stereo}, which consists in comparing the first two highest scores
\begin{equation}
   \text{Lowe's Ratio} = \frac{\text{Highest similarity score}}{\text{Second highest similarity score}}.
\end{equation}
A large  ratio means that the maximum score is significantly higher than the second maximum. In other words, the correspondence has a high level of confidence.

\begin{figure}[t]
    \centering
\begin{tblr}{hlines, vlines,
             colspec={*{3}{c}},
             }
{} & Auto-Encoder & Triplet Model      \\
\rotatebox{90}{\hspace{-2em} Orientation} & \includegraphics[height=.85in, width=1.3in, valign=m]{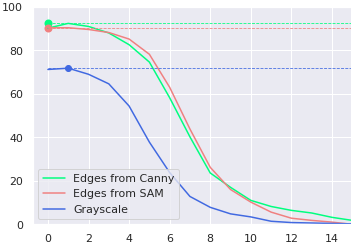} 
        &   \includegraphics[height=.85in, width=1.3in, valign=m]{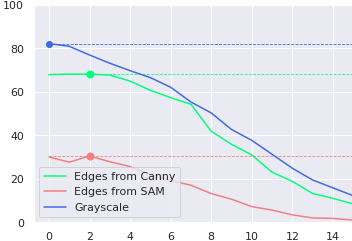}\\
\rotatebox{90}{\hspace{-1.5em} Altitude} & \includegraphics[height=.85in, width=1.3in, valign=m]{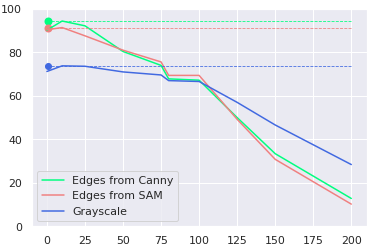}
        &   \includegraphics[height=.85in, width=1.3in, valign=m]{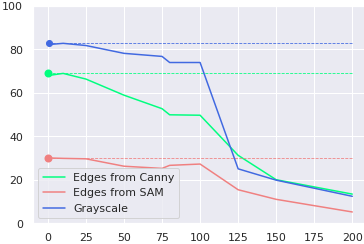}\\
\end{tblr}
\caption{\small Sensitivity to rotation and altitude drop for Auto-Encoder and Triplet Model, on trajectory B. For the row \textit{Orientation}, figures show location accuracy at 15m VS angle of variation (degrees). For the row \textit{Altitude}, figures show location accuracy at 15m VS altitude drop (meters).}
\label{fig:sensitivity}
    \end{figure}

\section{Results analysis}


For evaluating localization performance of a given trajectory, we determine for each frame associated with the UAV view whether the computed location is within 15m of the ground truth. For example, for an accuracy of 75\% at 15m: means that 75\% of the UAV's observations were correctly located within 15m of the ground truth. The choice of the radius at 15m is arbitrary, increasing the radius leads to higher figures. 

From the results (Table~\ref{table:results}) we see that for grayscale images,  BoVW  performs significantly worse than Auto-Encoder and Triplet, while Triplet seems to outperform Auto-Encoder at high altitudes.
When using edges the Auto-Encoder achieves the best results. Triplet Model doesn't seem to benefit from edge maps compared to grayscale images. Thus, we can recommend using the Auto-Encoder with  edge maps, 
moreover since the results with Canny and SAM are similar we prefer the lightweight Canny.

\smallskip
\noindent \textbf{Orientation sensitivity} In the previous experiment, the inference images are oriented in the same direction as the training ones. However, in practice, the orientation of the UAV can change (e.g. due to wind). Here, we investigate the robustness of each method when the orientation of the UAV is different from the orientation of the training images. We carry out inferences by voluntarily altering the orientation of the UAV images from 1 to 15 degrees. For each degree of alteration and each trajectory, the localization accuracy at 15m is computed.
Figure~\ref{fig:sensitivity} shows, for the Auto-Encoder method, that using edges provides both better localization when UAV rotation is introduced. (+40.4 \% accuracy using SAM-extracted contours rather than natural images for trajectory B, with UAV rotation of 5 degrees). As shown on the right in Figure~\ref{fig:sensitivity}, the Triplet Model does not appear to benefit from edge extraction.

\smallskip

\noindent \textbf{Altitude sensitivity} As in the case of orientation, the altitude of the drone can vary with respect to the planned trajectory. Thus, we also evaluated the robustness to changes in altitude. For that  we crop the UAV's view to simulate altitude drops down to -200m. 
Figure~\ref{fig:sensitivity} shows that using contours rather than grayscale images makes the Auto-Encoder method more robust altitude drops (up to -100m for a trajectory at 2000m altitude). 
Again, the Triplet Model doesn't seem to benefit particularly from the use of contours.


\smallskip 

\noindent {\bf Confidence criterion} 
In Table~\ref{table:results_lowe} are reported the results of the Auto-Encoder method applying Lowe's confidence criterion. The threshold for the criterion was arbitrarily chosen to $1.13$, so predictions with score below the threshold are not considered in the accuracy computation. Comparing with Table~\ref{table:results} we see that,  the confident matches have overall a higher accuracy. Since the low altitude trajectory (A) has a small field of view, many images are flat and ambiguous hence the higher rejection rate.  

\section{Conclusion}
In this work, we have demonstrated that using a edge extraction method upstream
of a UAV localization method based on Auto-Encoder results in significantly
better localization performance than state-of-the-art methods. In addition, we
show that it enhances robustness to a changes in UAV orientation and altitude drops, while also being invariant to seasonality. Using Canny,
combined with the Auto-Encoder, allows the method to be deployed onboard the
UAV and used in real time. Last, we have shown that using a confidence criterion
we can filter out most of the incorrect predictions.  Future research will
focus on  exploiting the causality of the localization along the trajectory.


{\small
\bibliographystyle{IEEEbib}
\bibliography{egbib}
}

\end{document}